\documentclass[10pt,twocolumn,letterpaper]{article}

\usepackage[pagenumbers]{cvpr} 

\definecolor{cvprblue}{rgb}{0.21,0.49,0.74}
\usepackage[pagebackref,breaklinks,colorlinks,allcolors=cvprblue]{hyperref}
\usepackage{graphicx}
\usepackage{epstopdf}
 \usepackage{multirow}
\usepackage{pifont}
\def\paperID{9891} 
\def\confName{CVPR}
\def\confYear{2025}

\title{Event USKT : U-State Space Model in Knowledge Transfer for Event Cameras}

\author{
    Yuhui Lin$^{1*}$\hspace{15mm}
    Jiahao Zhang$^1$\hspace{15mm}
    Siyuan Li$^2$\hspace{15mm}
    Jimin Xiao$^1$ \\[2mm]
    Ding Xu$^{3}$\hspace{15mm}
    Wenjun Wu$^4$\hspace{15mm}
    Jiaxuan Lu$^{5\dag}$ \\[4mm] 
    $^1$Xi'an Jiaotong-Liverpool University \\
    $^2$Dalian University of Technology \\
    $^3$Alibaba International Digital Business Group \\
    $^4$University of Illinois at Urbana-Champaign \\
    $^5$Shanghai Artificial Intelligence Laboratory\\[3mm]
    *First Author: \texttt{Yuhui.Lin21@student.xjtlu.edu.cn} \\ 
    \dag Corresponding Author: \texttt{lujiaxuan@pjlab.org.cn}
}

\begin{document}
\maketitle
\begin{abstract}
Event cameras, as an emerging imaging technology, offer distinct advantages over traditional RGB cameras, including reduced energy consumption and higher frame rates. However, the limited quantity of available event data presents a significant challenge, hindering their broader development. To alleviate this issue, we introduce a tailored U-shaped State Space Model Knowledge Transfer (USKT) framework for Event-to-RGB knowledge transfer. This framework generates inputs compatible with RGB frames, enabling event data to effectively reuse pre-trained RGB models and achieve competitive performance with minimal parameter tuning. Within the USKT architecture, we also propose a bidirectional reverse state space model. Unlike conventional bidirectional scanning mechanisms, the proposed Bidirectional Reverse State Space Model (BiR-SSM) leverages a shared weight strategy, which facilitates efficient modeling while conserving computational resources. In terms of effectiveness, integrating USKT with ResNet50 as the backbone improves model performance by 0.95\%, 3.57\%, and 2.9\% on DVS128 Gesture, N-Caltech101, and CIFAR-10-DVS datasets, respectively, underscoring USKT's adaptability and effectiveness. The code will be made available upon acceptance.
\end{abstract}    
\section{Introduction}

\begin{figure}[t]
\centering
\includegraphics[width=0.48\textwidth]{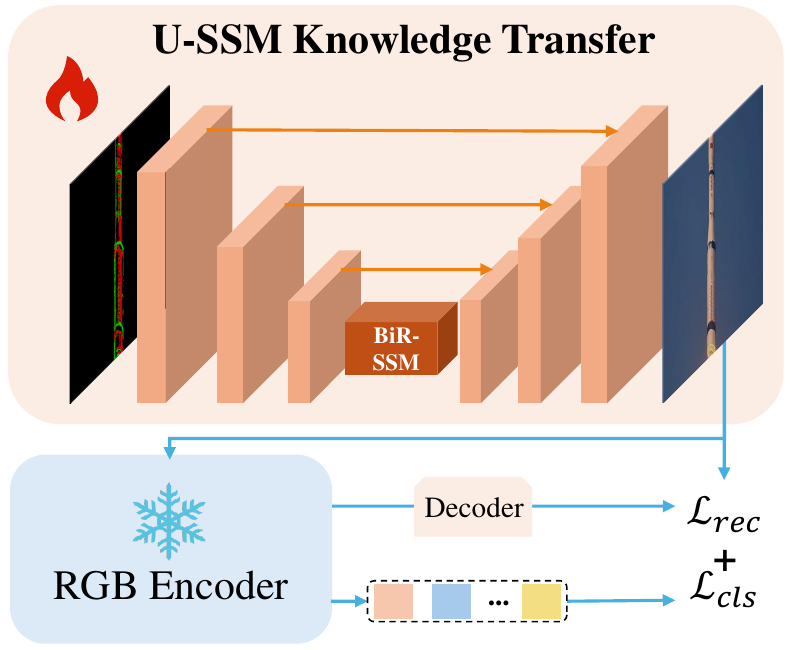} 
\caption{The proposed U-shaped State Space Model Knowledge Transfer (USKT) framework with the BiR-SSM module combines reconstruction and classification losses for Event-to-RGB feature adaptation, enabling the reuse of the pre-trained RGB encoder.} 
\label{fig:intro} 
\end{figure}


Event cameras represent a novel imaging technology that differs fundamentally from traditional frame-based cameras by capturing changes in brightness at the pixel level continuously, rather than capturing entire frames at regular intervals. The unique mechanism provides event cameras with exceptionally high temporal resolution and minimal latency, making them particularly well-suited for capturing fast-moving activities and handling scenes with high dynamic range \cite{rebecq2019high, gallego2020event}. Compared to traditional cameras, event cameras excel in environments with significant lighting variations, while also consuming less energy, making them highly promising for applications such as autonomous driving \cite{brebion2021real}, robotic navigation \cite{mueggler2018continuous}, and high-speed motion capture \cite{gao2023action}. However, as a relatively new imaging modality, event cameras face significant challenges related to data scarcity \cite{cadena2021spade, pan2020high}.

To address the challenge of limited data availability in event-based imaging, exploring knowledge transfer for event data is a promising direction worth investigating. In the broader field of knowledge transfer, methods can generally be categorized into domain-based and generative-based approaches. Domain-based methods aim to improve target domain performance by transferring knowledge from auxiliary domains \cite{zhang2021selective, moreno2012talmud, hu2018conet, lu2024pathotune}, while generative-based methods focus on generating synthetic data to enhance model performance \cite{tian2021knowledge, wang2020minegan, tan2020kt}.


In the field of event cameras, data is recorded only during changes in pixel brightness, resulting in event streams that are often sparse in visual content and differ significantly from the feature distributions of RGB images. Consequently, domain-based methods frequently encounter challenges related to domain mismatches, making effective knowledge transfer difficult \cite{kang2023event}. On the other hand, generative-based models can simulate sparse event streams to generate additional synthetic RGB data, which can be leveraged to enhance model training \cite{mostafavi2021e2sri, pan2020high}. 

To address the scarcity of event data, we design a generative U-shaped State Space Model Knowledge Transfer (USKT) framework tailored to the characteristics of event data. Previous research has widely recognized U-shaped methods for their excellent reconstruction capabilities \cite{esser2018variational, xie2023electrical}. Building upon these capabilities, we propose a generative knowledge transfer approach specifically for adapting event data to RGB features. As shown in Figure \ref{fig:intro}, our proposed method includes a Residual Down Sampling Block, a Residual Up Sampling Block, and a Bidirectional Reverse State Space Model. Specifically, the first Residual Down Sampling Block increases feature dimensionality while reducing spatial resolution, whereas the Residual Up Sampling Block enhances image restoration and preserves critical feature information, aligning the feature distribution more closely with that of RGB features. 

Furthermore, since convolutional in our model predominantly focus on local features during the downsampling process, we incorporate sequence modeling that captures global feature dependencies. As past Transformer-based approaches often faced significant computational resource demands \cite{he2023fourier, ren2023dlformer}, we introduce the Bidirectional Reverse State Space Model (BiR-SSM), which performs feature propagation through bidirectional scanning. Compared to the traditional Bidirectional State Space Model (Bi-SSM) \cite{zhu2024vision}, our BiR-SSM employs a shared SSM layer strategy aimed at ensuring feature consistency and reducing computational overhead.
Additionally, our approach simultaneously performs reconstruction and classification to improve the model's performance in classifying event images. In summary, our research makes the following three key contributions:
\begin{itemize}
    \item We introduce Event USKT, the first generative framework for knowledge transfer that adapts event data to pre-trained RGB models, establishing a new benchmark in this domain.
    \item We propose the Bidirectional Reverse State Space Model (BiR-SSM), which efficiently reduces computational overhead while ensuring effective feature adaptation.
    \item We present a hybrid loss function that synergistically combines reconstruction and classification objectives, significantly enhancing the performance of knowledge transfer for event image recognition.
\end{itemize}

\section{Related Work}\
\subsection{Event-based Image Recognition}

Event image recognition predominantly include graph-based models, Spiking Neural Networks (SNNs), and attention mechanisms. Graph-based models, using vertex and edge structures along with heterogeneous graph models and voxel grids, emulate spatial and temporal relationships among events and analyze complex data patterns, as demonstrated in various studies \cite{li2021graph, deng2022voxel, wang2023time, xie2022vmv, wu2020denoising, yuan2023learning, lu2023learning}. Spiking Neural Networks (SNNs) excel in processing time-step sequences for event image classification and, when integrated with attention mechanisms, significantly improve object recognition in dynamic environments by managing asynchronous data and focusing on critical features \cite{zhengGoingDeeperDirectlyTrained2020, fangIncorporatingLearnableMembrane2021a, zhaoBackEISNNDeepSpiking2022, fengMultiLevelFiringSpiking2023, duanTemporalEffectiveBatch, yao2021temporal, zhou2023bio, sironi2018hats, zhouspikformerwhenspiking2022}. Additionally, some attention-based methods have also been widely used \cite{liang2021event, deng2021mvf, li2023multi, jia2023event, gao2024hypergraph}. 



Among these methods, while tailored for event cameras, fail to address data scarcity. As a result, some studies have shifted to training with RGB-based models to mitigate this issue. For instance, several approaches based on ResNet have utilized RGB information to enhance the representational capability of event data \cite{klenk2024masked, deng2020amae}, while other methods have employed pre-trained ViT models based on RGB to improve the handling of sparse event streams \cite{wang2022exploiting}. Additionally, methods that integrate RGB and event camera data have proven to noticeably enhance the performance of downstream tasks \cite{yang2023event}.

\subsection{Knowledge Transfer}
In knowledge transfer, most approaches focus on RGB-to-RGB transfer. Domain adaptation methods align feature distributions between source and target domains. The PMC method enhances cross-modal recognition by generating missing target domain modalities through multimodal collaboration \cite{zhang2021progressive}. CLDA and MAJA mitigate domain shift via adversarial learning, boosting classification accuracy, especially in unsupervised scenarios \cite{he2020clda, zuo2021margin}. The DARDR method enhances cross-domain recognition by applying cross-modal constraints to transfer RGB-D data to the RGB target domain \cite{li2017domain}.

Generative-based methods use Generative Adversarial Networks (GANs) to generate target domain samples, reducing inter-domain differences. TriGAN and MSAN generate target samples from multiple source domains, significantly enhancing classification accuracy in unlabeled target domain tasks \cite{roy2021trigan, chen2020multiple}. DINE achieves privacy-preserving knowledge transfer with only a black-box source model \cite{liang2022dine}, while DupGAN employs a dual-GAN structure to effectively ensure feature consistency across domains \cite{hu2018duplex}. Meanwhile, U-shaped methods has shown promising results in generative tasks \cite{esser2018variational, xie2023electrical}.

\begin{figure*}
    \centering
    \includegraphics[width=1\linewidth]{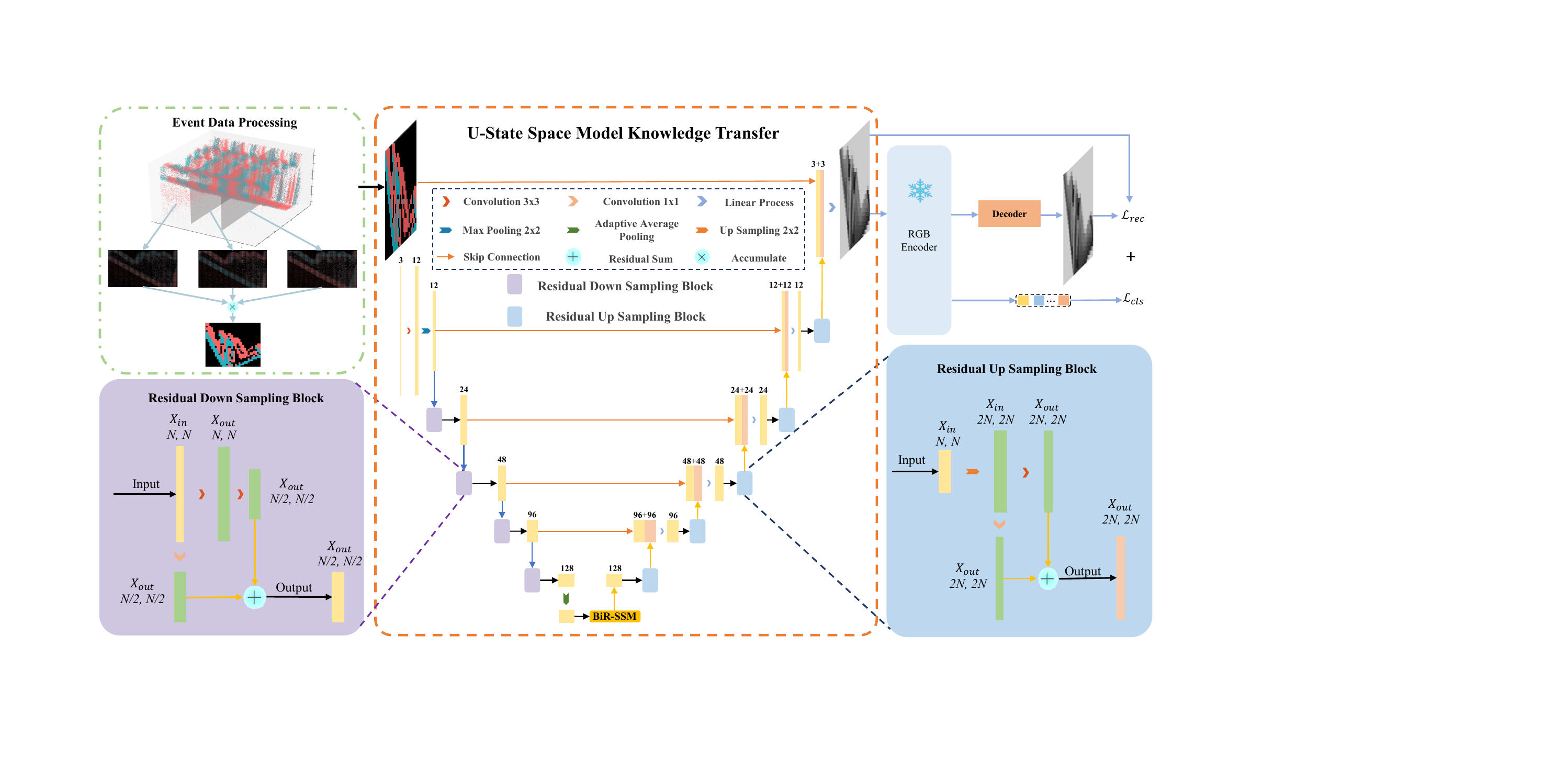}
    \caption{Overview of USKT framework. The proposed method is based on a U-shaped network, starting by mapping event data into suitable channels for USKT input through a time-accumulation. Subsequently, the data dimension is increased and the size is reduced via a downsampling process. Furthermore, we design a Bidirectional Reverse State Space Model (BiR-SSM) for sequence modeling. Following this, data is restored to its original resolution through an upsampling process. Finally, a reconstruction loss is introduced to enhance classification accuracy.
}
    \label{fig:pipeline}
\end{figure*}

In the research on knowledge transfer between Event and RGB modalities, a few studies adopt co-training approaches, where models integrate features from both modalities to enhance robustness and accuracy across various environments \cite{wang2024event, tomy2022fusing, li2024object}. In addition, CTN \cite{zhao2022transformer} is a Transformer-based cross-domain adaptation method that enhances the classification performance of event data by transferring features from RGB data.

\section{Method}

\subsection{Overview}

We propose a U-shaped State Space Model Knowledge Transfe (USKT) framework that efficiently converts event data into RGB features. As shown in Fig \ref{fig:pipeline}, the model consists of three key components: event data processing that transforms multiple time steps into voxel information to capture dynamic data changes; a Residual Down Sampling Block that reduces sequence length for efficient feature extraction; and a Residual Up Sampling Block that reconstructs the features into RGB domain suitable for encoder inputs. Additionally, we introduce a Bidirectional Reverse State Space Model (BiR-SSM) to fully capture the sequential dependencies between features. Finally, we focus on the performance of the Residual Up Sampling Block's output \(X_{\text{USKT}}\) after it passes through the feature extractor and present the design of the hybrid loss function.

\subsection{Event Data Processing}

An event stream can be visualized as consisting of multiple events, each characterized by \((x, y, t, p)\), where \((x, y)\) represent spatial coordinates, \(t\) denotes the timestamp, and \(p\) indicates the polarity (\(+1\) or \(-1\), signifying an increase or decrease in brightness). Consequently, event data is mapped into a three-dimensional grid where \((x, y)\) serve as the spatial dimensions and the time dimension is segmented into discrete bins, effectively organizing the event data temporally. Furthermore, based on the \((x, y)\) coordinates and the calculated time bin \(k\), the event polarity \(p\) is accumulated in the respective voxel within the grid. Each voxel \((x, y, k)\) then holds the aggregated polarity of events within the corresponding time bin, where the aggregation method, whether summing or averaging, depends on specific use cases. Ultimately, the final result is a three-dimensional tensor that retains both spatial and temporal information from the event stream.




\subsection{Generative U-SSM Knowledge Transfer}
Generative-based methods have been widely applied in the field of knowledge transfer across various tasks \cite{sohn2023visual, bai2019adaptive, yamaguchi2022transfer}, and U-Net-based architectures have shown promising results in generative tasks \cite{esser2018variational, xie2023electrical}. Building on these advances, we propose the U-SSM Knowledge Transfer (USKT) block for event-to-RGB knowledge transfer. Specifically, we input the event data $\mathbf{X}_{\text{input}} \in \mathbb{R}^{T \times 224 \times 224}$, where T represents the time steps. Using a convolutional layer, we map the input data to 12 dimensions, standardizing the time steps of the event camera. The convolution operation can be expressed as:


\begin{equation}
\mathbf{X}_{\text{proj}} = \mathbf{Conv}(\mathbf{X}_{\text{input}}),
\end{equation}


U-shaped models are highly effective for knowledge transfer, primarily due to the essential roles of their downsampling and upsampling modules. Downsampling modules compress data by reducing feature sizes and increasing dimensionality \cite{lu2023robust, he2023slicesamp}, whereas upsampling modules expand features and retain detailed information necessary for reconstruction \cite{dai2021learning, song2022inverted}. However, traditional U-shaped approaches, typically designed for RGB data, may not directly translate to event data, which primarily captures changes in brightness. The mismatch can lead to overfitting. Moreover, the inherent sparsity of event data necessitates a departure from conventional downsampling techniques; therefore, we incorporate residual connections to maintain the integrity of the original features. To address these challenges, we introduce the Residual Down Sampling Block and Residual Up Sampling Block for effective downsampling and upsampling, respectively. As illustrated in Figure~\ref{fig:pipeline}, the proposed framework employs 4 Residual Down Sampling Blocks and 5 Residual Up Sampling Blocks.


\noindent \textbf{Residual Down Sampling Block.} For the Block, the input feature $\mathbf{X}_{\text{proj}} \in \mathbb{R}^{D \times N \times N}$ undergoes a series of operations. First, a convolution operation is applied to extract global features, resulting in $\mathbf{X}_{\text{conv1}} \in \mathbb{R}^{F \times N \times N}$. Next, another convolution focuses on feature downsampling, producing $\mathbf{X}_{\text{conv2}} \in \mathbb{R}^{F \times N/2 \times N/2}$. Simultaneously, the original input feature is downsampled directly through a convolution layer, yielding $\mathbf{X}_{\text{res}} \in \mathbb{R}^{F \times N/2 \times N/2}$. A residual connection is then applied, resulting in $\mathbf{X}_{\text{down}} \in \mathbb{R}^{F \times N/2 \times N/2}$. The Residual Down Sampling Block preserves essential features while reducing the spatial dimensions of the data.
 



Meanwhile, in our method, the input is $\mathbf{X}_{\text{proj}} \in \mathbb{R}^{12 \times 224 \times 224}$ and the outpput is $\mathbf{X}_{\text{down}} \in \mathbb{R}^{128 \times 14 \times 14}$ sequentially. For the output of the final Residual Down Sampling Block, we apply an average pooling strategy to further reduce the spatial dimensions and computational complexity while preserving global information.


After downsampling process, our model employs BiR-SSM for feature modeling, which achieves effective feature propagation under relatively low computational resources. It will be further detailed in Section 3.4. 

\noindent \textbf{Residual Up Sampling Block.} For the Block, the input $\mathbf{X}_{\text{input}}$ is $\in \mathbb{R}^{D \times N \times N}$. Initially, we employ bilinear interpolation to enlarge the input dimensions. Subsequent feature extraction is performed using a $3 \times 3$ convolutional kernel followed by a $1 \times 1$ point convolution kernel. Afterwards, $\mathbf{X}_{\text{up}}$ is concatenated with the corresponding scale feature $\mathbf{X}_{\text{down}}$. To finalize the process, a convolutional fusion technique is applied to reduce the dimensionality.




Meanwhile, in our method, the input of the first block, we use the modeling result from BiR-SSM, \(\mathbf{X}_{\text{ssm}} \in \mathbb{R}^{128 \times 7 \times 7}\). And the result is \(\mathbf{X}_{\text{up}} \in \mathbb{R}^{D \times 224 \times 224}\) (\(D\) represents the dimension of the input provided to the USKT.) for each step. Additionally, if the final \(\mathbf{X}_{\text{up}}\) does not have a dimension of 3, we apply a convolution to the final \(\mathbf{X}_{\text{up}}\) to produce an output with the desired dimensions, \(\mathbf{X}_{\text{USKT}} \in \mathbb{R}^{3 \times 224 \times 224}\).

\subsection{Bidirectional Reverse State Space Model}

\begin{figure}[t]
\centering
\includegraphics[width=0.48\textwidth]{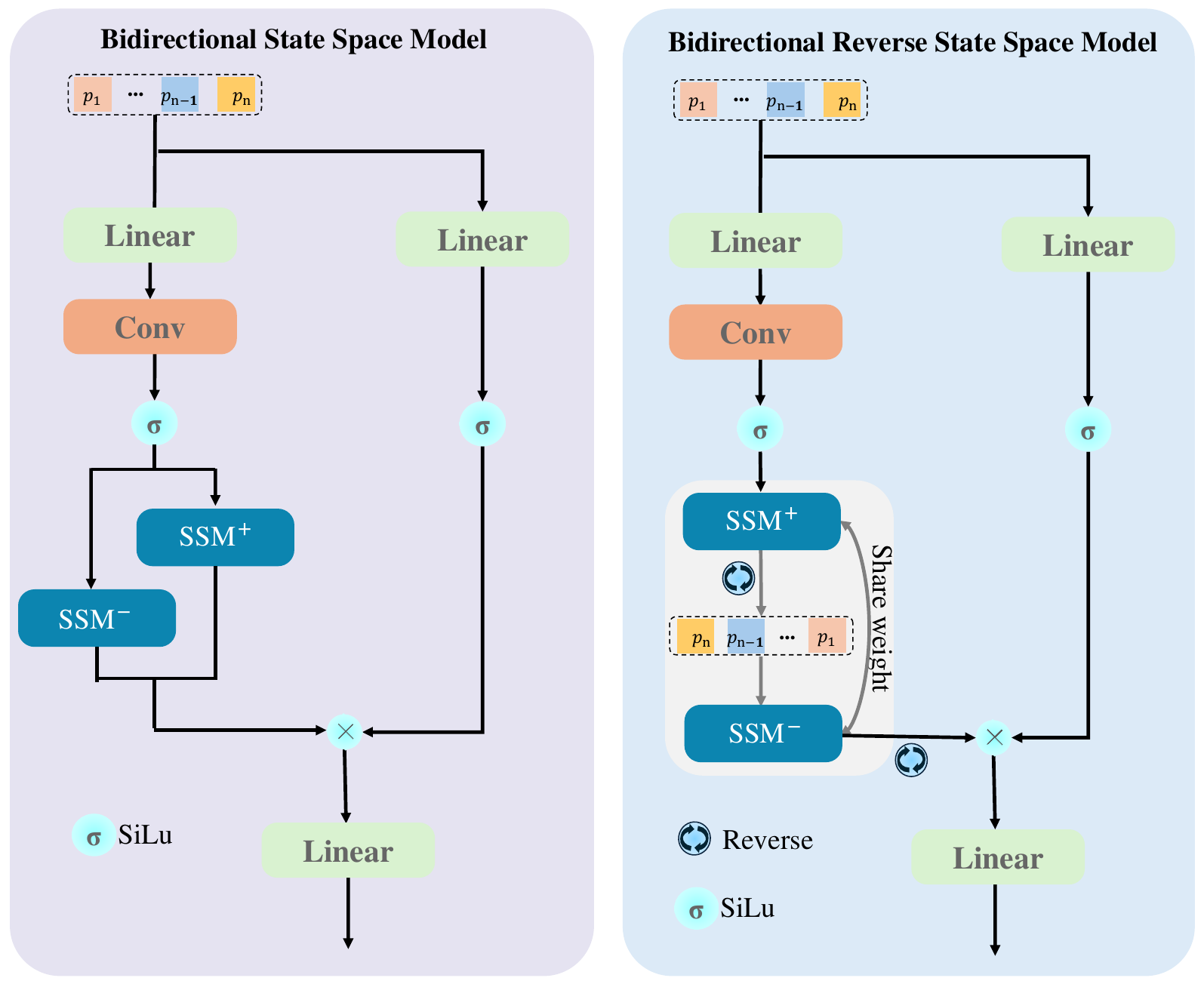}
\caption{The figure on the left shows the traditional Bi-SSM, while the figure on the right represents our proposed BiR-SSM.} 
\label{fig:ssm} 
\end{figure}

In this section, we focus on the application of a bidirectional reverse state space model for sequence modeling, as illustrated in Figure \ref{fig:ssm}. While Transformer-based methods offer substantial benefits for sequence modeling, their quadratic computational complexity often limits their performance \cite{he2023fourier, ren2023dlformer}. To address this, we conduct sequence modeling after the Residual Down Sampling Block, which enables efficient processing while preserving essential feature information. Notably, previous bidirectional state space models typically relied on two separate SSM layers \cite{zhu2024vision}, as depicted in Figure \ref{fig:ssm}. We believe this design can be optimized to improve parameter efficiency without sacrificing model performance.


After the Residual Down Sampling Block, we flatten the 2D data into a 1D sequence and then employ the Bidirectional Reverse State Space Model to process the downsampled data, represented as \( X_{\text{down}} \). The sequence is then processed through a linear layer, a convolutional layer, and a State Space Model (SSM) layer. To retain original information, the output from the SSM undergoes a residual connection with the original sequence. The downsampled data \( X_{\text{down}} \) is represented as a set of features \(\{ p_1, p_2, \dots, p_n \}\) where each \( p_n \) is an element of \( X_{\text{down}} \), as shown in the following formula:
\begin{equation}
    X_{Conv} = \mathit{Conv}(\mathit{Linear}(X_{\text{down}})),
\end{equation}
where \(X_{Conv}\) is obtained after passing through a linear layer and a convolutional mapping.

Simultaneously, when we obtain the result of \( \mathit{SSM^+} \), we apply a SiLU function to the output of \( \mathit{SSM^+} \) after processing through \( X_{Conv} \), as shown in the following formula:
\begin{equation}
    X_{SSM^+} = \mathit{SiLU}(\mathit{SSM^+}(X_{Conv})),
\end{equation}
where the forward \(\mathit{SSM^+}\) modeling is applied to the features to obtain \(X_{SSM^+}\). 

The result of \(\mathit{SSM^+}\) is then reversed, represented as a set of features \(\{ p_n, \dots, p_2, p_1 \}\) where each \( p_1 \) is an element from \( X_{\mathit{SSM^-}} \), to facilitate the subsequent \( \mathit{SSM^-} \) processing. Finally, the sequence passes through \( \mathit{SSM^-} \), as shown in the following formula:
\begin{equation}
    X_{\mathit{SSM^-}} = \mathit{SiLU}(\mathit{SSM^-}(X_{\mathit{SSM^+}})),
\end{equation}
where \(\mathit{SSM^-}\) modeling is applied to the features.

We apply a residual connection between \( X_{\mathit{SSM^-}} \) and \( X_{\mathit{down}} \), producing \( X \). Then, we reverse the resulting sequence to restore the original structural arrangement, \(\{ p_1, p_2, \dots, p_n \}\) where each \( p_n \) is an element of \(\mathit{X_{\mathit{SSM}}}\), the reversed form of \( X \).

\begin{equation}
    X = \mathit{Linear}(X_{\mathit{SSM^-}} + \mathit{SiLU}(\mathit{Linear}(\mathit{X_{\mathit{down}}}))),
\end{equation}
where a residual connection is used to combine \(X_{\mathit{SSM^-}}\) and \(\mathit{X_{\mathit{down}}}\) to mitigate feature loss.

\subsection{Reconstruction and Classification}
\textbf{Feature Extraction.} We use the ResNet \cite{he2016deep} as pre-trained RGB Encoder for feature extraction. Unlike traditional ResNet applications, we utilize the adaptive output from USKT as the input for Encoder. After feature extraction through Encoder, the feature matrix $\mathbf{X}_{\text{res}} \in \mathbb{R}^{D \times 7 \times 7}$. Subsequently, as shown in Fig \ref{fig:pipeline} through decoder, the features are mapped back to the original space, ultimately resulting in an output  $\mathbf{X}_{\text{rec}} \in \mathbb{R}^{3 \times 224 \times 224}$. In our method, our decoder employs a deconvolution approach.



\noindent \textbf{Loss Function.} In our proposed method, we primarily used two types of loss functions. For the classification, we applied the Focal Loss to the classification results from the linear layer, is defined as:

\begin{equation}
    \mathcal{L}_{\text{cls}} = -\alpha_t (1 - p_t)^\gamma \log(p_t),
\end{equation}
where \(p_t\) is the predicted probability for the correct class \(t\), \(\alpha_t\) balances positive and negative samples, and \(\gamma\) focuses on hard-to-classify sample.

For the reconstruction part, we use the Mean Squared Error (MSE) loss function to compare the reconstructed features \(\mathbf{X}_{\text{rec}} \in \mathbb{R}^{3 \times 224 \times 224}\) and  \(\mathbf{X}_{\text{USKT}} \in \mathbb{R}^{3 \times 224 \times 224}\), is defined as:

\begin{equation}
    \mathcal{L}_{\text{rec}} = \frac{1}{N} \sum_{i=1}^{N} \left( \mathbf{X}_{\text{rec}}^{(i)} - \mathbf{X}_{\text{USKT}}^{(i)} \right)^2,
\end{equation}
where \(N\) is the total number of elements, and \(i\) indexes the elements.

Finally, we combine \(\lambda_1\) and \(\lambda_2\) to compute our total loss \(\mathcal{L}\), defined as:
\begin{equation}
    \mathcal{L} = \lambda_1 \cdot L_{\text{cls}} + \lambda_2 \cdot L_{\text{rec}},
\end{equation}
where  \(\lambda_1\) and \(\lambda_2\) are the weights for the classification loss and reconstruction loss, respectively.

\section{Experiments}
\subsection{Experimental Setup}

\begin{table*}[t]
\centering
\caption{Comparison of classification accuracies on the DVS128 Gesture, N-Caltech101, and CIFAR-10-DVS, showing the top-1 accuracy.}
\begin{tabular}{@{}lccclccc@{}}
\toprule
Method     & KT & Supervised & Pre-training   & Architecture & DVS128 Gesture & N-Caltech101 & CIFAR-10-DVS \\ \midrule
ViT \cite{dosovitskiy2020image}     & \ding{55} & \checkmark    & \ding{55}          & ViT-S/16     & 56.06  & 55.63        & 52.45        \\
ResNet \cite{he2016deep}   & \ding{55} & \checkmark    & \ding{55}          & ResNet50     & 68.65  & 62.69        & 56.65        \\ 
ViT \cite{dosovitskiy2020image}    & \ding{55}  & \checkmark    & \checkmark & ViT-S/16  & 68.18  & 85.02        & 76.10        \\ \midrule

SimCLR \cite{chen2020simple}   & \ding{55} & \ding{55} & \checkmark & ResNet50 & 84.47& 86.57        & 75.15        \\ 
MoCo-v2 \cite{chen2020improved}  & \ding{55} & \ding{55} & \checkmark & ResNet50 & 87.12  & 84.16        & 74.65        \\
MoCo-v3 \cite{chen2021empirical}   & \ding{55} & \ding{55} & \checkmark & ViT-S/16 & 64.39  & 76.59        & 68.40        \\ \midrule

STBP-tdBN\cite{zhengGoingDeeperDirectlyTrained2020} & \ding{55} & \checkmark& \ding{55}& SNN& 78.94& 66.01&66.93\\
PLIF\cite{fangIncorporatingLearnableMembrane2021a} & \ding{55} & \checkmark& \ding{55}& SNN& 79.98& 64.73&66.77\\
MLF\cite{fengMultiLevelFiringSpiking2023} & \ding{55} & \checkmark& \ding{55}& SNN& 85.77& 70.42&67.07\\
Spikformer\cite{li2022spikeformer} & \ding{55} & \checkmark& \ding{55}& SNN& 79.52& 72.83&68.55\\ \midrule

PKOA\cite{he2019knowledge} & \checkmark & \checkmark & \checkmark & ResNet50 & 85.07& 86.65&76.35\\
CAF\cite{xie2022collaborative} & \checkmark & \checkmark & \checkmark& ResNet101& 86.81& 85.98&75.30\\
PMC\cite{zhang2021progressive} & \checkmark & \ding{55} & \checkmark& ResNet18& 77.51& 78.70& 70.15\\
DINE\cite{liang2022dine} & \checkmark & \ding{55} & \checkmark & ResNet50& 79.25& 80.52& 70.00\\
CTN \cite{zhao2022transformer}  & \checkmark & \ding{55} & \checkmark & ResNet50 & 88.10  & 90.10    & 77.30   \\
MAJA\cite{zuo2021margin} & \checkmark & \ding{55} & \checkmark& ResNet101& 82.72& 84.77& 72.35\\
TriGAN\cite{roy2021trigan} & \checkmark & \ding{55} & \checkmark& ResNet101& 83.76 & 82.95& 75.20\\ \midrule
Ours(Frozen)     & \checkmark  & \checkmark   & \checkmark           & ResNet50  & 85.23  & 88.82        & 76.75        \\
Ours(Unfrozen)    & \checkmark   & \checkmark      & \checkmark           & ResNet50  & \textbf{89.02}  & \textbf{90.35}        & \textbf{79.10}        \\ \bottomrule
\end{tabular}
\end{table*}



\paragraph{Dataset.}
We utilize the ImageNet-1K dataset \cite{deng2009imagenet} for pre-training our models. In our experiments, we compare SimCLR, MoCo-v2, and MoCo-v3, all pretrained on both ImageNet-1K \cite{deng2009imagenet} and N-ImageNet \cite{kim2021n}. Furthermore, we extend our knowledge transfer activities to the DVS128 Gesture \cite{amir2017low}, N-Caltech101 \cite{orchard2015converting}, and CIFAR-10-DVS \cite{cheng2020structure} datasets to assess the generalization capabilities of our models across various domains. Additionally, we adapt the input by resizing images to a resolution of 224×224 pixels.

DVS128 Gesture \cite{amir2017low} consists of 1,188 event streams from 29 participants, categorized into 11 gesture types, with each event stream featuring a resolution of approximately 128×128 pixels. N-Caltech101 \cite{orchard2015converting} comprises a total of 8,242 images across 101 categories, with each image having a resolution of around 300×200 pixels. CIFAR-10-DVS \cite{cheng2020structure} includes 10 classes, with 1,000 samples per class, totaling 10,000 samples, each at a resolution of 128×128 pixels.

\paragraph{Implementation.}
Our model is implemented using PyTorch and trained on NVIDIA RTX 2080Ti GPUs. For all experiments, we employ the AdamW optimizer \cite{loshchilov2017decoupled} and utilize a cosine scheduler. The initial learning rate is set to 0.0025, with a reduced rate of 0.000025 for fine-tuning layers.

\subsection{Comparison with Existing Methods}

\paragraph{Compared to RGB-based Supervised Methods.}

In the non-pretrained models, our method achieved significant improvements on the DVS128 Gesture, N-Caltech101, and CIFAR-10-DVS datasets compared to VIT-S/16 and ResNet50. Specifically, our method outperformed VIT-S/16 by 29.17\%, 33.19\%, and 24.3\%, and surpassed ResNet50 by 16.58\%, 26.13\%, and 20.1\% respectively. For the pre-trained models, our method outperformed VIT-S/16 by 17.05\% on DVS128 Gesture, 3.80\% on N-Caltech101, and 0.6\% on CIFAR-10-DVS. It demonstrates that our experiments show significant performance improvements under both pretrained and non-pretrained supervised conditions.

\paragraph{Compared to RGB-based Unsupervised Methods.}
We used a frozen ResNet50 backbone to compare with traditional RGB unsupervised methods.
On the DVS128 Gesture, our frozen model can outperform many unsupervised models, surpassing SimCLR \cite{chen2020simple} and MoCo-v3 \cite{chen2021empirical} by 0.76\% and 2.65\%, respectively. On the N-Caltech101, our model also outperforms many unsupervised models, surpassing SimCLR \cite{chen2020simple} and MoCo-v2 \cite{chen2020improved} by 2.25\% and 4.66\%, respectively. On the CIFAR-10-DVS, our model also surpasses many unsupervised models, exceeding SimCLR \cite{chen2020simple} and MoCo-v2 \cite{chen2020improved} by 1.6\% and 2.1\%, respectively.
In the unfrozen condition, on the DVS128 Gesture, our model can surpass MoCo-v2 \cite{chen2020improved} by 2.10\%. Therefore, compared to traditional RGB unsupervised methods, our model demonstrates significant advantages.

\paragraph{Compared to SNN methods.}
Due to the effective handling of event information by SNNs in event camera classification tasks, our method(frozen) was compared with SNN-based methods. On the DVS128 Gesture, our model showed significant advantages over other advanced SNN-based methods, not only outperforming Spikformer \cite{li2022spikeformer} by 5.71\% but also achieving a comparable level to MLF \cite{fengMultiLevelFiringSpiking2023}. Similarly, on the N-Caltech101, our model excelled, surpassing Spikformer \cite{li2022spikeformer} by 15.99\%. Furthermore, on the CIFAR-10-DVS, our model further demonstrated its superiority, outperforming Spikformer \cite{li2022spikeformer}, MLF \cite{fengMultiLevelFiringSpiking2023}, and TEBN \cite{duanTemporalEffectiveBatch} by 8.2\%, 9.68\%, and 9.05\%, respectively. These results fully demonstrate the excellent performance and leading position of our model in handling tasks based on SNNs.

\begin{table*}[t]
\centering
\caption{Comparison of the performance of ResNet18, ResNet34, and ResNet50 with and without the implementation of USKT, illustrating top-1 accuracy on the DVS128 Gesture, N-Caltech101, and CIFAR-10-DVS datasets. The table also delineates the results under both frozen and unfrozen backbone conditions.}
\label{tab:comparison_resnet_backbone}
\begin{tabular}{@{}lcccccccc@{}}
\toprule
Method    & USKT & Frozen & Params & GFlops & DVS128 Gesture  & N-Caltech101 & CIFAR-10-DVS \\ \midrule
\multirow{4}{*}{Resnet18}  & \ding{55}   & \checkmark        & 0.01M & 1.82GMac & 80.25 & 81.63 & 69.05\\ 
                           & \ding{55}   & \ding{55}         & 11.18M & 1.82GMac & 82.77 & 83.87 & 73.75\\
                           & \checkmark  & \checkmark        & 7.63M & 3.44GMac & 83.19 (\textbf{+2.94})  & 83.85 (\textbf{+2.22}) & 72.50 (\textbf{+3.45})\\ 
                           & \checkmark  & \ding{55}         & 18.53M & 3.44GMac & 84.45 (\textbf{+1.68}) & 85.07 (\textbf{+1.20}) & 75.90 (\textbf{+2.15})\\ \midrule
\multirow{4}{*}{Resnet34}  & \ding{55}   & \checkmark        & 0.014M & 3.68GMac & 82.63 & 83.75 & 72.95\\
                           & \ding{55}   & \ding{55}         & 21.29M & 3.68GMac & 84.45 & 87.49 & 76.80\\
                           & \checkmark  & \checkmark        & 7.37M & 5.29GMac & 84.87 (\textbf{+2.24}) & 85.55 (\textbf{+1.80}) & 75.05 (\textbf{+2.10})\\
                           & \checkmark  & \ding{55}         & 28.64M & 5.29GMac & 86.70 (\textbf{+2.25}) & 87.98 (\textbf{+0.49}) & 77.95 (\textbf{+1.15})\\ 
\midrule
\multirow{4}{*}{Resnet50}  & \ding{55}   & \checkmark        & 0.05M & 4.13GMac & 84.28 & 85.25 & 73.85\\
                           & \ding{55}   & \ding{55}         & 23.53M & 4.13GMac & 86.55 & 87.77 & 77.10\\
                           & \checkmark  & \checkmark        & 10.71M & 5.90GMac & 85.23 (\textbf{+0.95}) & 88.82 (\textbf{+3.57}) & 76.75 (\textbf{+2.90}) \\
                           & \checkmark  & \ding{55}         & 34.19M & 5.90GMac & 89.02 (\textbf{+2.47}) & 90.35 (\textbf{+2.58}) & 79.10 (\textbf{+2.00})\\ 
\bottomrule
\end{tabular}
\end{table*}

\paragraph{Compared to Knowledge Transfer Methods.}
We primarily demonstrate the superiority of our method by comparing it with knowledge transfer-based approaches. In supervised methods, our model with a frozen backbone network can train with very low parameter counts, surpassing PKOA \cite{he2019knowledge} by 1.17\% on N-Caltech101 and by 0.4\% on CIFAR-10-DVS. When the backbone network is unfrozen, our model further exceeds PKOA \cite{he2019knowledge} by 3.7\% on N-Caltech101 and by 2.75\% on CIFAR-10-DVS, and surpasses CAF \cite{xie2022collaborative} by 3.21\% on DVS128-Gesture.

In unsupervised methods, our proposed method with a frozen backbone outperforms TriGAN \cite{roy2021trigan} by 1.47\%, 5.87\%, and 1.55\% on DVS128-Gesture, N-Caltech101, and CIFAR-10-DVS, respectively. Unfreezing the backbone allows our model to further exceed CTN \cite{zhao2022transformer} by 0.92\%, 0.25\%, and 1.8\% on DVS128-Gesture, N-Caltech101, and CIFAR-10-DVS, respectively.

\subsection{Abaltion Studies}

In this section, we address three key issues: Firstly, we examine the applicability of our proposed USKT Block to various sizes of ResNet models. Secondly, we assess the effectiveness of the USKT Block in enhancing model performance. Thirdly, we conduct a comparative analysis of the BiR-SSM Block.

\paragraph{Adaptability of USKT.}

As shown in Table~\ref{tab:comparison_resnet_backbone}, we evaluated the performance of our proposed USKT across different sizes of ResNet on the DVS128 Gesture, N-Caltech101, and CIFAR-10-DVS datasets to validate the applicability of USKT to various ResNet architectures.

Initially, we conducted experiments with the backbone network frozen (with only the bias parameters of ResNet unfrozen). Using ResNet18 as the backbone, the integration of USKT resulted in performance improvements of 2.94\%, 2.12\%, and 3.45\% on the DVS128 Gesture, N-Caltech101, and CIFAR-10-DVS datasets, respectively. With ResNet34 as the backbone, USKT enhanced the model's performance by 2.24\%, 1.8\%, and 2.1\% on these respective datasets. When employing ResNet50, the addition of USKT led to gains of 0.95\%, 3.57\%, and 2.9\%.



Further, we evaluated the performance on the N-Caltech101 with the backbone network completely unfrozen. In this dataset, adding USKT improved the model's performance by 1.2\% with ResNet18, 0.49\% with ResNet34, and 2.58\% with ResNet50 as the backbone.

Table~\ref{tab:comparison_resnet_backbone} illustrates that our method achieves the most substantial improvements with the ResNet50 backbone, irrespective of the network's state (frozen or unfrozen). Its superior performance is likely attributable to ResNet50's enhanced capability to extract richer fine-grained information from images compared to the ResNet18 and ResNet34 models.

\paragraph{Effectiveness of USKT.}
\begin{table}[t]
\centering
\caption{Comparison of different domain-adaptive generation methods for classification accuracies, showing top-1 accuracy on the N-Caltech101.}
\label{tab:comparison_object_recognition}
\begin{tabular}{@{}lccc@{}}
\toprule
Method       & Frozen & Params   & N-Caltech101  \\ \midrule
Conv-1       & \checkmark    & 8.80    & 86.22 \\ 
Conv-2       & \checkmark    & 8.80    & 85.88 \\ 
U-Conv+Trans & \checkmark    & 11.83   & 86.68 \\
U-Conv+Trans & \ding{55}   & 35.31   & 88.20 \\ \midrule
USKT         & \checkmark    & 10.71   & 88.82 \\
USKT         & \ding{55}     & 34.19   & \textbf{90.35} \\ \midrule
\end{tabular}
\end{table}


As illustrated in Table~\ref{tab:comparison_object_recognition}, we conducted comparative evaluations between convolution and Transformer-based methods to validate the effectiveness of our proposed USKT. Initially, we substituted USKT with convolutional layers to assess the adaptive capabilities of our approach. The experiments were executed with a frozen ResNet50 backbone. Our model demonstrated improvements of 2.6\% and 2.94\% over single and double convolution layer setups, respectively.

Furthermore, to rigorously assess the efficacy of our proposed BiR-SSM, we carried out comparative experiments under both frozen and unfrozen conditions of the ResNet50 backbone, where BiR-SSM was replaced with a Transformer module. Under the frozen condition, our method exceeded the performance of the Transformer-based methods by 2.14\%, achieving results comparable to those of the unfrozen backbone Transformer. Remarkably, even with fewer parameters in the unfrozen state, our approach not only matched but surpassed the Transformer-based model by 2.15\%.

\paragraph{Comparision of BiR-SSM Block.}
\begin{table}[t]
\centering
\caption{Comparison of different ssm layers for classification accuracies, showing top-1 accuracy on N-Caltech101.}
\label{tab:comparison_BiR}
\begin{tabular}{@{}lcc@{}}
\toprule
Method       & Layer Number    & N-Caltech101 \\ \midrule
SSM-forward  &  2     & 87.19 \\ 
SSM-backward &  2     & 86.78 \\ 
Bi-SSM & 1   & 87.68 \\ \midrule
BiR-SSM (ours) &  1 & \textbf{88.20} \\ \midrule
\end{tabular}
\end{table}
As shown in Table~\ref{tab:comparison_BiR}, we have frozen the ResNet50 backbone and substituted the original SSM with our novel BiR-SSM in various configurations. It can be concluded that our proposed BiR-SSM outperforms the traditional Bi-SSM. This enhancement is likely attributable to the improved data consistency achieved through the shared SSM mechanism that we implemented.

\subsection{Hyperparameter Studies}
This section first discusses the impact of different numbers of BiR-SSM layers on the model, followed by an analysis of different \(\lambda_2\) affect model performance.


\begin{figure}[t]
\centering
\includegraphics[width=0.48\textwidth]{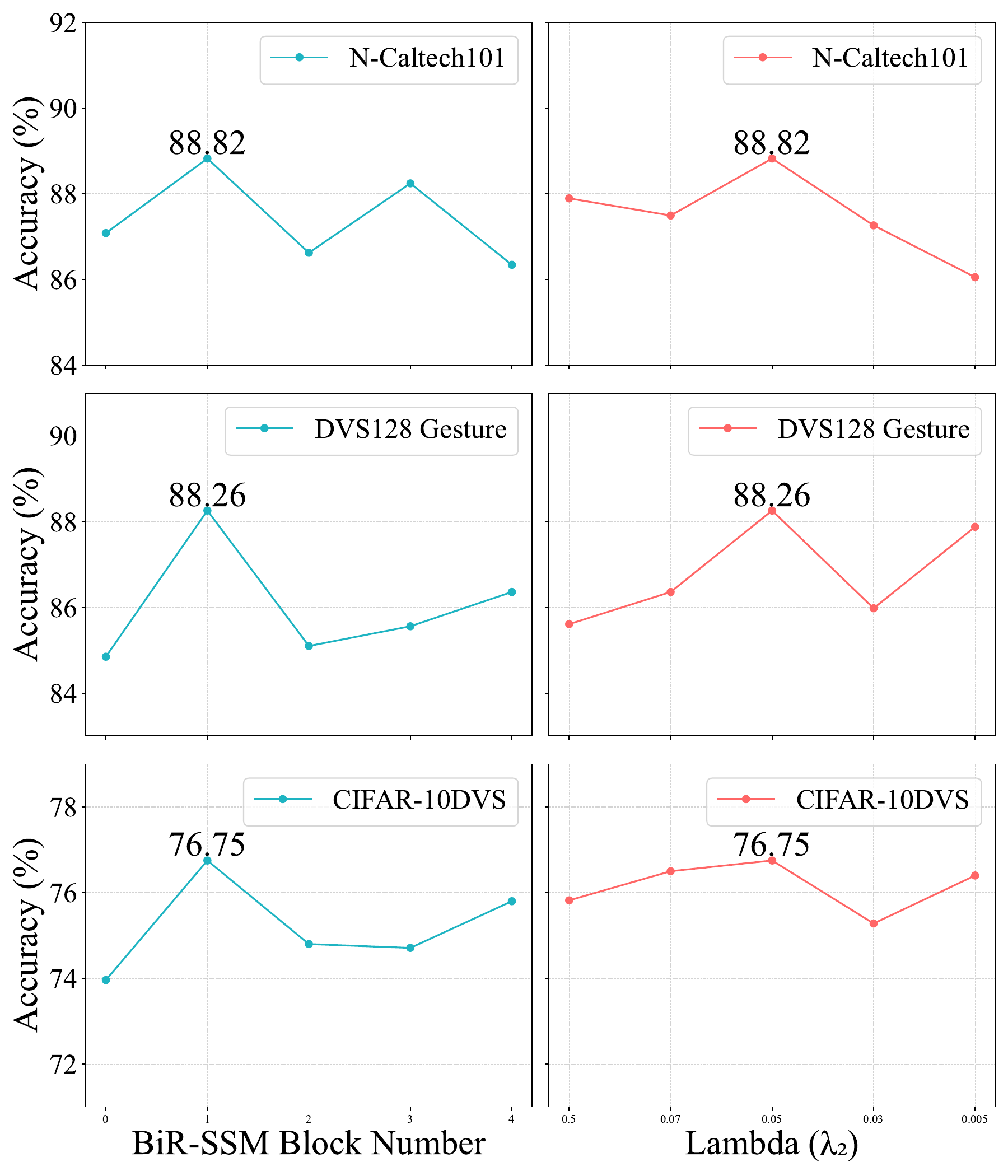} 
\caption{The left is the comparison of the performance of ResNet50 with different numbers of SSM layers in USKT and the right is the comparison of the performance of ResNet50 with different \(\lambda_2\), showing top-1 accuracy on DVS128 Gesture, N-Caltech101 and CIAFR-10-DVS.} 
\label{fig:ssmhy} 
\end{figure}


\paragraph{Comparison of Different Number of BiR-SSM Layers.}
As demonstrated in Figure~\ref{fig:ssmhy}, employing a single SSM layer yields the highest accuracy, surpassing the configurations where no BiR-SSM layers or multiple BiR-SSM layers are used. In our setup, we utilized ResNet-50 as the backbone with the main network components frozen. Specifically, with one BiR-SSM layer, our method achieved an accuracy of 88.82\% on the N-Caltech101 dataset. Similarly, this configuration attained accuracies of 88.62\% on the DVS128 Gesture and 76.75\% on CIFAR-10-DVS. We hypothesize that the absence of any BiR-SSM layers causes the adaptive domain to predominantly focus on local information, thereby neglecting global context. Conversely, incorporating more than one BiR-SSM layer can lead to overfitting or an excessive emphasis on the classification task, potentially compromising the model's performance.




\paragraph{Comparison of Different \(\lambda_2\) Settings.}

As depicted in Figure~\ref{fig:ssmhy}, varying the parameter \(\lambda_2\) significantly influences the performance of our model. In our evaluations, we employed a frozen ResNet-50 architecture as the backbone, and set \(\lambda_1\) to 1. The results indicate optimal performance when \(\lambda_2\) is set to 0.05, with the model achieving an accuracy of 88.82\% on the N-Caltech101 dataset. Similar efficacy is observed on the DVS128 Gesture with an accuracy of 88.62\%, and a noteworthy performance of 76.75\% on CIFAR-10-DVS. We hypothesize that a \(\lambda_2\) value below 0.05 potentially leads the model to prioritize the classification task, possibly at the expense of generalization capabilities. Conversely, a \(\lambda_2\) value above 0.05 seems to excessively focus the model on the reconstruction task, which detrimentally impacts classification accuracy.

\section{Conclusion}
In this paper, we introduce the USKT framework to tackle the challenge of limited event data in event-based imaging by facilitating effective Event-to-RGB knowledge transfer. The framework allows event data to leverage pre-trained RGB models with minimal tuning, achieving robust performance. Our BiR-SSM component, with its shared weight strategy, further enhances computational efficiency. Experimental results across multiple datasets demonstrate USKT’s adaptability and effectiveness in advancing event-based imaging.
{
    \small
    \bibliographystyle{ieeenat_fullname}
    \bibliography{main}

\begin{thebibliography}{79}
\providecommand{\natexlab}[1]{#1}
\providecommand{\url}[1]{\texttt{#1}}
\expandafter\ifx\csname urlstyle\endcsname\relax
  \providecommand{\doi}[1]{doi: #1}\else
  \providecommand{\doi}{doi: \begingroup \urlstyle{rm}\Url}\fi

\bibitem[Amir et~al.(2017)Amir, Taba, Berg, Melano, McKinstry, Di~Nolfo, Nayak, Andreopoulos, Garreau, Mendoza, et~al.]{amir2017low}
Arnon Amir, Brian Taba, David Berg, Timothy Melano, Jeffrey McKinstry, Carmelo Di~Nolfo, Tapan Nayak, Alexander Andreopoulos, Guillaume Garreau, Marcela Mendoza, et~al.
\newblock A low power, fully event-based gesture recognition system.
\newblock In \emph{Proceedings of the IEEE Conference on Computer Vision and Pattern Recognition}, pages 7243--7252, 2017.

\bibitem[Bai et~al.(2019)Bai, Quan, and Luo]{bai2019adaptive}
Wenjun Bai, Changqin Quan, and Zhi-Wei Luo.
\newblock Adaptive generative initialization in transfer learning.
\newblock \emph{Computer and Information Science 17}, pages 63--74, 2019.

\bibitem[Brebion et~al.(2021)Brebion, Moreau, and Davoine]{brebion2021real}
Vincent Brebion, Julien Moreau, and Franck Davoine.
\newblock Real-time optical flow for vehicular perception with low-and high-resolution event cameras.
\newblock \emph{IEEE Transactions on Intelligent Transportation Systems}, 23\penalty0 (9):\penalty0 15066--15078, 2021.

\bibitem[Cadena et~al.(2021)Cadena, Qian, Wang, and Yang]{cadena2021spade}
Pablo Rodrigo~Gantier Cadena, Yeqiang Qian, Chunxiang Wang, and Ming Yang.
\newblock Spade-e2vid: Spatially-adaptive denormalization for event-based video reconstruction.
\newblock \emph{IEEE Transactions on Image Processing}, 30:\penalty0 2488--2500, 2021.

\bibitem[Chen et~al.(2020{\natexlab{a}})Chen, Xie, Wen, Huang, and Ding]{chen2020multiple}
Chaoqi Chen, Weiping Xie, Yi Wen, Yue Huang, and Xinghao Ding.
\newblock Multiple-source domain adaptation with generative adversarial nets.
\newblock \emph{Knowledge-Based Systems}, 199:\penalty0 105962, 2020{\natexlab{a}}.

\bibitem[Chen et~al.(2020{\natexlab{b}})Chen, Kornblith, Norouzi, and Hinton]{chen2020simple}
Ting Chen, Simon Kornblith, Mohammad Norouzi, and Geoffrey Hinton.
\newblock A simple framework for contrastive learning of visual representations.
\newblock In \emph{International Conference on Machine Learning}, pages 1597--1607, 2020{\natexlab{b}}.

\bibitem[Chen et~al.(2020{\natexlab{c}})Chen, Fan, Girshick, and He]{chen2020improved}
Xinlei Chen, Haoqi Fan, Ross Girshick, and Kaiming He.
\newblock Improved baselines with momentum contrastive learning.
\newblock \emph{arXiv Preprint arXiv:2003.04297}, 2020{\natexlab{c}}.

\bibitem[Chen et~al.(2021)Chen, Xie, and He]{chen2021empirical}
Xinlei Chen, Saining Xie, and Kaiming He.
\newblock An empirical study of training self-supervised vision transformers.
\newblock In \emph{Proceedings of the IEEE/CVF International Conference on Computer Vision}, pages 9640--9649, 2021.

\bibitem[Cheng et~al.(2020)Cheng, Luo, Yang, Yu, and Li]{cheng2020structure}
Wensheng Cheng, Hao Luo, Wen Yang, Lei Yu, and Wei Li.
\newblock Structure-aware network for lane marker extraction with dynamic vision sensor.
\newblock \emph{arXiv Preprint arXiv:2008.06204}, 2020.

\bibitem[Dai et~al.(2021)Dai, Lu, and Shen]{dai2021learning}
Yutong Dai, Hao Lu, and Chunhua Shen.
\newblock Learning affinity-aware upsampling for deep image matting.
\newblock In \emph{Proceedings of the IEEE/CVF Conference on Computer Vision and Pattern Recognition}, pages 6841--6850, 2021.

\bibitem[Deng et~al.(2009)Deng, Dong, Socher, Li, Li, and Fei-Fei]{deng2009imagenet}
Jia Deng, Wei Dong, Richard Socher, Li-Jia Li, Kai Li, and Li Fei-Fei.
\newblock Imagenet: A large-scale hierarchical image database.
\newblock In \emph{2009 IEEE Conference on Computer Vision and Pattern Recognition}, pages 248--255, 2009.

\bibitem[Deng et~al.(2020)Deng, Li, and Chen]{deng2020amae}
Yongjian Deng, Youfu Li, and Hao Chen.
\newblock Amae: Adaptive motion-agnostic encoder for event-based object classification.
\newblock \emph{IEEE Robotics and Automation Letters}, 5\penalty0 (3):\penalty0 4596--4603, 2020.

\bibitem[Deng et~al.(2021)Deng, Chen, and Li]{deng2021mvf}
Yongjian Deng, Hao Chen, and Youfu Li.
\newblock Mvf-net: A multi-view fusion network for event-based object classification.
\newblock \emph{IEEE Transactions on Circuits and Systems for Video Technology}, 32\penalty0 (12):\penalty0 8275--8284, 2021.

\bibitem[Deng et~al.(2022)Deng, Chen, Liu, and Li]{deng2022voxel}
Yongjian Deng, Hao Chen, Hai Liu, and Youfu Li.
\newblock A voxel graph cnn for object classification with event cameras.
\newblock In \emph{Proceedings of the IEEE/CVF Conference on Computer Vision and Pattern Recognition}, pages 1172--1181, 2022.

\bibitem[Dosovitskiy(2020)]{dosovitskiy2020image}
Alexey Dosovitskiy.
\newblock An image is worth 16x16 words: Transformers for image recognition at scale.
\newblock \emph{arXiv Preprint arXiv:2010.11929}, 2020.

\bibitem[Duan et~al.()Duan, Ding, Chen, Yu, and Huang]{duanTemporalEffectiveBatch}
Chaoteng Duan, Jianhao Ding, Shiyan Chen, Zhaofei Yu, and Tiejun Huang.
\newblock Temporal effective batch normalization in spiking neural networks.

\bibitem[Esser et~al.(2018)Esser, Sutter, and Ommer]{esser2018variational}
Patrick Esser, Ekaterina Sutter, and Bj{\"o}rn Ommer.
\newblock A variational u-net for conditional appearance and shape generation.
\newblock In \emph{Proceedings of the IEEE Conference on Computer Vision and Pattern Recognition}, pages 8857--8866, 2018.

\bibitem[Fang et~al.(2021)Fang, Yu, Chen, Masquelier, Huang, and Tian]{fangIncorporatingLearnableMembrane2021a}
Wei Fang, Zhaofei Yu, Yanqi Chen, Timothee Masquelier, Tiejun Huang, and Yonghong Tian.
\newblock Incorporating learnable membrane time constant to enhance learning of spiking neural networks, 2021.

\bibitem[Feng et~al.(2023)Feng, Liu, Tang, Ma, and Pan]{fengMultiLevelFiringSpiking2023}
Lang Feng, Qianhui Liu, Huajin Tang, De Ma, and Gang Pan.
\newblock Multi-level firing with spiking ds-resnet: Enabling better and deeper directly-trained spiking neural networks, 2023.

\bibitem[Gallego et~al.(2020)Gallego, Delbr{\"u}ck, Orchard, Bartolozzi, Taba, Censi, Leutenegger, Davison, Conradt, Daniilidis, et~al.]{gallego2020event}
Guillermo Gallego, Tobi Delbr{\"u}ck, Garrick Orchard, Chiara Bartolozzi, Brian Taba, Andrea Censi, Stefan Leutenegger, Andrew~J Davison, J{\"o}rg Conradt, Kostas Daniilidis, et~al.
\newblock Event-based vision: A survey.
\newblock \emph{IEEE Transactions on Pattern Analysis and Machine Intelligence}, 44\penalty0 (1):\penalty0 154--180, 2020.

\bibitem[Gao et~al.(2023)Gao, Lu, Li, Ma, Du, Li, and Dai]{gao2023action}
Yue Gao, Jiaxuan Lu, Siqi Li, Nan Ma, Shaoyi Du, Yipeng Li, and Qionghai Dai.
\newblock Action recognition and benchmark using event cameras.
\newblock \emph{IEEE Transactions on Pattern Analysis and Machine Intelligence}, 2023.

\bibitem[Gao et~al.(2024)Gao, Lu, Li, Li, and Du]{gao2024hypergraph}
Yue Gao, Jiaxuan Lu, Siqi Li, Yipeng Li, and Shaoyi Du.
\newblock Hypergraph-based multi-view action recognition using event cameras.
\newblock \emph{IEEE Transactions on Pattern Analysis and Machine Intelligence}, 2024.

\bibitem[He et~al.(2016)He, Zhang, Ren, and Sun]{he2016deep}
Kaiming He, Xiangyu Zhang, Shaoqing Ren, and Jian Sun.
\newblock Deep residual learning for image recognition.
\newblock In \emph{Proceedings of the IEEE Conference on Computer Vision and Pattern Recognition}, pages 770--778, 2016.

\bibitem[He and Wang(2023)]{he2023slicesamp}
Lianlian He and Ming Wang.
\newblock Slicesamp: A promising downsampling alternative for retaining information in a neural network.
\newblock \emph{Applied Sciences}, 13\penalty0 (21):\penalty0 11657, 2023.

\bibitem[He et~al.(2019)He, Shen, Tian, Gong, Sun, and Yan]{he2019knowledge}
Tong He, Chunhua Shen, Zhi Tian, Dong Gong, Changming Sun, and Youliang Yan.
\newblock Knowledge adaptation for efficient semantic segmentation.
\newblock In \emph{Proceedings of the IEEE/CVF Conference on Computer Vision and Pattern Recognition}, pages 578--587, 2019.

\bibitem[He et~al.(2020)He, Yang, Chen, Mu, and Li]{he2020clda}
Zhihai He, Bo Yang, Chaoxian Chen, Qilin Mu, and Zesong Li.
\newblock Clda: An adversarial unsupervised domain adaptation method with classifier-level adaptation.
\newblock \emph{Multimedia Tools and Applications}, 79:\penalty0 33973--33991, 2020.

\bibitem[He et~al.(2023)He, Yang, Feng, Yin, Wang, Leng, and Lin]{he2023fourier}
Ziwei He, Meng Yang, Minwei Feng, Jingcheng Yin, Xinbing Wang, Jingwen Leng, and Zhouhan Lin.
\newblock Fourier transformer: Fast long range modeling by removing sequence redundancy with fft operator.
\newblock \emph{arXiv Preprint arXiv:2305.15099}, 2023.

\bibitem[Hu et~al.(2018{\natexlab{a}})Hu, Zhang, and Yang]{hu2018conet}
Guangneng Hu, Yu Zhang, and Qiang Yang.
\newblock Conet: Collaborative cross networks for cross-domain recommendation.
\newblock In \emph{Proceedings of the 27th ACM International Conference on Information and Knowledge Management}, pages 667--676, 2018{\natexlab{a}}.

\bibitem[Hu et~al.(2018{\natexlab{b}})Hu, Kan, Shan, and Chen]{hu2018duplex}
Lanqing Hu, Meina Kan, Shiguang Shan, and Xilin Chen.
\newblock Duplex generative adversarial network for unsupervised domain adaptation.
\newblock In \emph{Proceedings of the IEEE Conference on Computer Vision and Pattern Recognition}, pages 1498--1507, 2018{\natexlab{b}}.

\bibitem[Jia et~al.(2023)Jia, You, He, Tian, Feng, Wang, Jia, Lou, Zhang, Li, et~al.]{jia2023event}
Zexi Jia, Kaichao You, Weihua He, Yang Tian, Yongxiang Feng, Yaoyuan Wang, Xu Jia, Yihang Lou, Jingyi Zhang, Guoqi Li, et~al.
\newblock Event-based semantic segmentation with posterior attention.
\newblock \emph{IEEE Transactions on Image Processing}, 32:\penalty0 1829--1842, 2023.

\bibitem[Kang and Kang(2023)]{kang2023event}
Daehyun Kang and Dongwoo Kang.
\newblock Event camera-based pupil localization: Facilitating training with event-style translation of rgb faces.
\newblock \emph{IEEE Access}, 2023.

\bibitem[Kim et~al.(2021)Kim, Bae, Park, Zhang, and Kim]{kim2021n}
Junho Kim, Jaehyeok Bae, Gangin Park, Dongsu Zhang, and Young~Min Kim.
\newblock N-imagenet: Towards robust, fine-grained object recognition with event cameras.
\newblock In \emph{Proceedings of the IEEE/CVF International Conference on Computer Vision}, pages 2146--2156, 2021.

\bibitem[Klenk et~al.(2024)Klenk, Bonello, Koestler, Araslanov, and Cremers]{klenk2024masked}
Simon Klenk, David Bonello, Lukas Koestler, Nikita Araslanov, and Daniel Cremers.
\newblock Masked event modeling: Self-supervised pretraining for event cameras.
\newblock In \emph{Proceedings of the IEEE/CVF Winter Conference on Applications of Computer Vision}, pages 2378--2388, 2024.

\bibitem[Li and Liu(2023)]{li2023multi}
Lin Li and Yang Liu.
\newblock Multi-dimensional attention spiking transformer for event-based image classification.
\newblock pages 359--362, 2023.

\bibitem[Li et~al.(2024)Li, Linger, Millhaeusler, Tsiminaki, Li, and Dai]{li2024object}
Lei Li, Alexander Linger, Mario Millhaeusler, Vagia Tsiminaki, Yuanyou Li, and Dengxin Dai.
\newblock Object-centric cross-modal feature distillation for event-based object detection.
\newblock In \emph{2024 IEEE International Conference on Robotics and Automation (ICRA)}, pages 15440--15447, 2024.

\bibitem[Li et~al.(2017)Li, Fang, Zhang, and Wu]{li2017domain}
Xiao Li, Min Fang, Ju-Jie Zhang, and Jinqiao Wu.
\newblock Domain adaptation from rgb-d to rgb images.
\newblock \emph{Signal Processing}, 131:\penalty0 27--35, 2017.

\bibitem[Li et~al.(2021)Li, Zhou, Yang, Zhang, Cui, Bao, and Zhang]{li2021graph}
Yijin Li, Han Zhou, Bangbang Yang, Ye Zhang, Zhaopeng Cui, Hujun Bao, and Guofeng Zhang.
\newblock Graph-based asynchronous event processing for rapid object recognition.
\newblock In \emph{Proceedings of the IEEE/CVF International Conference on Computer Vision}, pages 934--943, 2021.

\bibitem[Li et~al.(2022)Li, Lei, and Yang]{li2022spikeformer}
Yudong Li, Yunlin Lei, and Xu Yang.
\newblock Spikeformer: A novel architecture for training high-performance low-latency spiking neural network.
\newblock \emph{arXiv Preprint arXiv:2211.10686}, 2022.

\bibitem[Liang et~al.(2022)Liang, Hu, Feng, and He]{liang2022dine}
Jian Liang, Dapeng Hu, Jiashi Feng, and Ran He.
\newblock Dine: Domain adaptation from single and multiple black-box predictors.
\newblock In \emph{Proceedings of the IEEE/CVF Conference on Computer Vision and Pattern Recognition}, pages 8003--8013, 2022.

\bibitem[Liang et~al.(2021)Liang, Chen, Li, Liu, and Knoll]{liang2021event}
Zichen Liang, Guang Chen, Zhijun Li, Peigen Liu, and Alois Knoll.
\newblock Event-based object detection with lightweight spatial attention mechanism.
\newblock In \emph{2021 6th IEEE International Conference on Advanced Robotics and Mechatronics (ICARM)}, pages 498--503, 2021.

\bibitem[Loshchilov(2017)]{loshchilov2017decoupled}
I Loshchilov.
\newblock Decoupled weight decay regularization.
\newblock \emph{arXiv Preprint arXiv:1711.05101}, 2017.

\bibitem[Lu et~al.(2024)Lu, Yan, Zhang, Gao, and Zhang]{lu2024pathotune}
Jiaxuan Lu, Fang Yan, Xiaofan Zhang, Yue Gao, and Shaoting Zhang.
\newblock Pathotune: Adapting visual foundation model to pathological specialists.
\newblock In \emph{International Conference on Medical Image Computing and Computer-Assisted Intervention}, pages 395--406, 2024.

\bibitem[Lu et~al.(2023{\natexlab{a}})Lu, Chen, Tang, Ding, and Luo]{lu2023robust}
Wei Lu, Si-Bao Chen, Jin Tang, Chris~HQ Ding, and Bin Luo.
\newblock A robust feature downsampling module for remote-sensing visual tasks.
\newblock \emph{IEEE Transactions on Geoscience and Remote Sensing}, 61:\penalty0 1--12, 2023{\natexlab{a}}.

\bibitem[Lu et~al.(2023{\natexlab{b}})Lu, Wang, Liu, Wang, and Wang]{lu2023learning}
Yunfan Lu, Zipeng Wang, Minjie Liu, Hongjian Wang, and Lin Wang.
\newblock Learning spatial-temporal implicit neural representations for event-guided video super-resolution.
\newblock pages 1557--1567, 2023{\natexlab{b}}.

\bibitem[Moreno et~al.(2012)Moreno, Shapira, Rokach, and Shani]{moreno2012talmud}
Orly Moreno, Bracha Shapira, Lior Rokach, and Guy Shani.
\newblock Talmud: Transfer learning for multiple domains.
\newblock In \emph{Proceedings of the 21st ACM International Conference on Information and Knowledge Management}, pages 425--434, 2012.

\bibitem[Mostafavi et~al.(2021)Mostafavi, Nam, Choi, and Yoon]{mostafavi2021e2sri}
Mohammad Mostafavi, Yeongwoo Nam, Jonghyun Choi, and Kuk-Jin Yoon.
\newblock E2sri: Learning to super-resolve intensity images from events.
\newblock \emph{IEEE Transactions on Pattern Analysis and Machine Intelligence}, 44\penalty0 (10):\penalty0 6890--6909, 2021.

\bibitem[Mueggler et~al.(2018)Mueggler, Gallego, Rebecq, and Scaramuzza]{mueggler2018continuous}
Elias Mueggler, Guillermo Gallego, Henri Rebecq, and Davide Scaramuzza.
\newblock Continuous-time visual-inertial odometry for event cameras.
\newblock \emph{IEEE Transactions on Robotics}, 34\penalty0 (6):\penalty0 1425--1440, 2018.

\bibitem[Orchard et~al.(2015)Orchard, Jayawant, Cohen, and Thakor]{orchard2015converting}
Garrick Orchard, Ajinkya Jayawant, Gregory~K Cohen, and Nitish Thakor.
\newblock Converting static image datasets to spiking neuromorphic datasets using saccades.
\newblock \emph{Frontiers in Neuroscience}, 9:\penalty0 437, 2015.

\bibitem[Pan et~al.(2020)Pan, Hartley, Scheerlinck, Liu, Yu, and Dai]{pan2020high}
Liyuan Pan, Richard Hartley, Cedric Scheerlinck, Miaomiao Liu, Xin Yu, and Yuchao Dai.
\newblock High frame rate video reconstruction based on an event camera.
\newblock \emph{IEEE Transactions on Pattern Analysis and Machine Intelligence}, 44\penalty0 (5):\penalty0 2519--2533, 2020.

\bibitem[Rebecq et~al.(2019)Rebecq, Ranftl, Koltun, and Scaramuzza]{rebecq2019high}
Henri Rebecq, Ren{\'e} Ranftl, Vladlen Koltun, and Davide Scaramuzza.
\newblock High speed and high dynamic range video with an event camera.
\newblock \emph{IEEE Transactions on Pattern Analysis and Machine Intelligence}, 43\penalty0 (6):\penalty0 1964--1980, 2019.

\bibitem[Ren et~al.(2023)Ren, Wang, and Huang]{ren2023dlformer}
Lei Ren, Haiteng Wang, and Gao Huang.
\newblock Dlformer: A dynamic length transformer-based network for efficient feature representation in remaining useful life prediction.
\newblock \emph{IEEE Transactions on Neural Networks and Learning Systems}, 2023.

\bibitem[Roy et~al.(2021)Roy, Siarohin, Sangineto, Sebe, and Ricci]{roy2021trigan}
Subhankar Roy, Aliaksandr Siarohin, Enver Sangineto, Nicu Sebe, and Elisa Ricci.
\newblock Trigan: Image-to-image translation for multi-source domain adaptation.
\newblock \emph{Machine Vision and Applications}, 32:\penalty0 1--12, 2021.

\bibitem[Sironi et~al.(2018)Sironi, Brambilla, Bourdis, Lagorce, and Benosman]{sironi2018hats}
Amos Sironi, Manuele Brambilla, Nicolas Bourdis, Xavier Lagorce, and Ryad Benosman.
\newblock Hats: Histograms of averaged time surfaces for robust event-based object classification.
\newblock In \emph{Proceedings of the IEEE Conference on Computer Vision and Pattern Recognition}, pages 1731--1740, 2018.

\bibitem[Sohn et~al.(2023)Sohn, Chang, Lezama, Polania, Zhang, Hao, Essa, and Jiang]{sohn2023visual}
Kihyuk Sohn, Huiwen Chang, Jos{\'e} Lezama, Luisa Polania, Han Zhang, Yuan Hao, Irfan Essa, and Lu Jiang.
\newblock Visual prompt tuning for generative transfer learning.
\newblock In \emph{Proceedings of the IEEE/CVF Conference on Computer Vision and Pattern Recognition}, pages 19840--19851, 2023.

\bibitem[Song et~al.(2022)Song, Zhao, Hui, Jiang, et~al.]{song2022inverted}
Zhaoyang Song, Xiaoqiang Zhao, Yongyong Hui, Hongmei Jiang, et~al.
\newblock Inverted n-type lightweight network based on back projection for image super-resolution reconstruction.
\newblock 2022.

\bibitem[Tan et~al.(2020)Tan, Liu, Liu, Yin, and Li]{tan2020kt}
Hongchen Tan, Xiuping Liu, Meng Liu, Baocai Yin, and Xin Li.
\newblock Kt-gan: Knowledge-transfer generative adversarial network for text-to-image synthesis.
\newblock \emph{IEEE Transactions on Image Processing}, 30:\penalty0 1275--1290, 2020.

\bibitem[Tian et~al.(2021)Tian, Zhang, Wang, Xiang, and Pan]{tian2021knowledge}
Kun Tian, Chenghao Zhang, Ying Wang, Shiming Xiang, and Chunhong Pan.
\newblock Knowledge mining and transferring for domain adaptive object detection.
\newblock In \emph{Proceedings of the IEEE/CVF International Conference on Computer Vision}, pages 9133--9142, 2021.

\bibitem[Tomy et~al.(2022)Tomy, Paigwar, Mann, Renzaglia, and Laugier]{tomy2022fusing}
Abhishek Tomy, Anshul Paigwar, Khushdeep~S Mann, Alessandro Renzaglia, and Christian Laugier.
\newblock Fusing event-based and rgb camera for robust object detection in adverse conditions.
\newblock In \emph{2022 International Conference on Robotics and Automation (ICRA)}, pages 933--939, 2022.

\bibitem[Wang et~al.(2023)Wang, Wang, and He]{wang2023time}
Ruilin Wang, Li Wang, and Yingbo He.
\newblock A time-related voxel representation method for event camera.
\newblock pages 553--557, 2023.

\bibitem[Wang et~al.(2024)Wang, Wang, Tang, Zhu, Jiang, Tian, and Tang]{wang2024event}
Xiao Wang, Shiao Wang, Chuanming Tang, Lin Zhu, Bo Jiang, Yonghong Tian, and Jin Tang.
\newblock Event stream-based visual object tracking: A high-resolution benchmark dataset and a novel baseline.
\newblock In \emph{Proceedings of the IEEE/CVF Conference on Computer Vision and Pattern Recognition}, pages 19248--19257, 2024.

\bibitem[Wang et~al.(2020)Wang, Gonzalez-Garcia, Berga, Herranz, Khan, and Weijer]{wang2020minegan}
Yaxing Wang, Abel Gonzalez-Garcia, David Berga, Luis Herranz, Fahad~Shahbaz Khan, and Joost van~de Weijer.
\newblock Minegan: Effective knowledge transfer from gans to target domains with few images.
\newblock In \emph{Proceedings of the IEEE/CVF Conference on Computer Vision and Pattern Recognition}, pages 9332--9341, 2020.

\bibitem[Wang et~al.(2022)Wang, Hu, and Liu]{wang2022exploiting}
Zuowen Wang, Yuhuang Hu, and Shih-Chii Liu.
\newblock Exploiting spatial sparsity for event cameras with visual transformers.
\newblock In \emph{2022 IEEE International Conference on Image Processing (ICIP)}, pages 411--415, 2022.

\bibitem[Wu et~al.(2020)Wu, Ma, Yu, and Shi]{wu2020denoising}
Jinjian Wu, Chuanwei Ma, Xiaojie Yu, and Guangming Shi.
\newblock Denoising of event-based sensors with spatial-temporal correlation.
\newblock pages 4437--4441, 2020.

\bibitem[Xie et~al.(2022{\natexlab{a}})Xie, Deng, Shao, Liu, and Li]{xie2022vmv}
Bochen Xie, Yongjian Deng, Zhanpeng Shao, Hai Liu, and Youfu Li.
\newblock Vmv-gcn: Volumetric multi-view based graph cnn for event stream classification.
\newblock \emph{IEEE Robotics and Automation Letters}, 7\penalty0 (2):\penalty0 1976--1983, 2022{\natexlab{a}}.

\bibitem[Xie et~al.(2022{\natexlab{b}})Xie, Li, Lv, Liu, Wang, and Wu]{xie2022collaborative}
Binhui Xie, Shuang Li, Fangrui Lv, Chi~Harold Liu, Guoren Wang, and Dapeng Wu.
\newblock A collaborative alignment framework of transferable knowledge extraction for unsupervised domain adaptation.
\newblock \emph{IEEE Transactions on Knowledge and Data Engineering}, 35\penalty0 (7):\penalty0 6518--6533, 2022{\natexlab{b}}.

\bibitem[Xie et~al.(2023)Xie, Fu, and Liu]{xie2023electrical}
Zhihui Xie, Min Fu, and Xuefeng Liu.
\newblock Electrical fittings inspection based on improved unet with generative adversarial network and attention mechanism.
\newblock In \emph{2023 8th International Conference on Image, Vision and Computing (ICIVC)}, pages 776--782, 2023.

\bibitem[Yamaguchi et~al.(2022)Yamaguchi, Kanai, Kumagai, Chijiwa, and Kashima]{yamaguchi2022transfer}
Shin'ya Yamaguchi, Sekitoshi Kanai, Atsutoshi Kumagai, Daiki Chijiwa, and Hisashi Kashima.
\newblock Transfer learning with pre-trained conditional generative models.
\newblock \emph{arXiv Preprint arXiv:2204.12833}, 2022.

\bibitem[Yang et~al.(2023)Yang, Pan, and Liu]{yang2023event}
Yan Yang, Liyuan Pan, and Liu Liu.
\newblock Event camera data pre-training.
\newblock In \emph{Proceedings of the IEEE/CVF International Conference on Computer Vision}, pages 10699--10709, 2023.

\bibitem[Yao et~al.(2021)Yao, Gao, Zhao, Wang, Lin, Yang, and Li]{yao2021temporal}
Man Yao, Huanhuan Gao, Guangshe Zhao, Dingheng Wang, Yihan Lin, Zhaoxu Yang, and Guoqi Li.
\newblock Temporal-wise attention spiking neural networks for event streams classification.
\newblock In \emph{Proceedings of the IEEE/CVF International Conference on Computer Vision}, pages 10221--10230, 2021.

\bibitem[Yuan et~al.(2023)Yuan, Jin, Wu, Wei, Wang, Chen, and Wang]{yuan2023learning}
Chengguo Yuan, Yu Jin, Zongzhen Wu, Fanting Wei, Yangzirui Wang, Lan Chen, and Xiao Wang.
\newblock Learning bottleneck transformer for event image-voxel feature fusion based classification.
\newblock pages 3--15, 2023.

\bibitem[Zhang et~al.(2021{\natexlab{a}})Zhang, Kong, and Zhang]{zhang2021selective}
Hongwei Zhang, Xiangwei Kong, and Yujia Zhang.
\newblock Selective knowledge transfer for cross-domain collaborative recommendation.
\newblock \emph{IEEE Access}, 9:\penalty0 48039--48051, 2021{\natexlab{a}}.

\bibitem[Zhang et~al.(2021{\natexlab{b}})Zhang, Xu, Zhang, and Ouyang]{zhang2021progressive}
Weichen Zhang, Dong Xu, Jing Zhang, and Wanli Ouyang.
\newblock Progressive modality cooperation for multi-modality domain adaptation.
\newblock \emph{IEEE Transactions on Image Processing}, 30:\penalty0 3293--3306, 2021{\natexlab{b}}.

\bibitem[Zhao et~al.(2022{\natexlab{a}})Zhao, Zeng, and Li]{zhaoBackEISNNDeepSpiking2022}
Dongcheng Zhao, Yi Zeng, and Yang Li.
\newblock Backeisnn: A deep spiking neural network with adaptive self-feedback and balanced excitatory--inhibitory neurons.
\newblock \emph{Neural Networks}, 154:\penalty0 68--77, 2022{\natexlab{a}}.

\bibitem[Zhao et~al.(2022{\natexlab{b}})Zhao, Zhang, and Huang]{zhao2022transformer}
Junwei Zhao, Shiliang Zhang, and Tiejun Huang.
\newblock Transformer-based domain adaptation for event data classification.
\newblock pages 4673--4677, 2022{\natexlab{b}}.

\bibitem[Zheng et~al.(2020)Zheng, Wu, Deng, Hu, and Li]{zhengGoingDeeperDirectlyTrained2020}
Hanle Zheng, Yujie Wu, Lei Deng, Yifan Hu, and Guoqi Li.
\newblock Going deeper with directly-trained larger spiking neural networks, 2020.

\bibitem[Zhou et~al.(2023)Zhou, Zheng, and Li]{zhou2023bio}
Qian Zhou, Peng Zheng, and Xiaohu Li.
\newblock A bio-inspired hierarchical spiking neural network with biological synaptic plasticity for event camera object recognition.
\newblock \emph{Sheng Wu Yi Xue Gong Cheng Xue Za Zhi= Journal of Biomedical Engineering= Shengwu Yixue Gongchengxue Zazhi}, 40\penalty0 (4):\penalty0 692--699, 2023.

\bibitem[Zhou et~al.(2022)Zhou, Zhu, He, Wang, Yan, Tian, and Yuan]{zhouspikformerwhenspiking2022}
Zhaokun Zhou, Yuesheng Zhu, Chao He, Yaowei Wang, Shuicheng Yan, Yonghong Tian, and Li Yuan.
\newblock Spikformer: When spiking neural network meets transformer, 2022.

\bibitem[Zhu et~al.(2024)Zhu, Liao, Zhang, Wang, Liu, and Wang]{zhu2024vision}
Lianghui Zhu, Bencheng Liao, Qian Zhang, Xinlong Wang, Wenyu Liu, and Xinggang Wang.
\newblock Vision mamba: Efficient visual representation learning with bidirectional state space model.
\newblock \emph{arXiv Preprint arXiv:2401.09417}, 2024.

\bibitem[Zuo et~al.(2021)Zuo, Yao, Zhuang, and Xu]{zuo2021margin}
Yukun Zuo, Hantao Yao, Liansheng Zhuang, and Changsheng Xu.
\newblock Margin-based adversarial joint alignment domain adaptation.
\newblock \emph{IEEE Transactions on Circuits and Systems for Video Technology}, 32\penalty0 (4):\penalty0 2057--2067, 2021.

\end{thebibliography}
}


\end{document}